\def\year{2018}\relax
\documentclass[letterpaper]{article} 
\usepackage{aaai18}
\usepackage{booktabs} 
\usepackage{graphicx} 
\usepackage{subfigure}
\usepackage{marvosym}

\usepackage[linesnumbered, algoruled]{algorithm2e}

\usepackage{enumitem}

\usepackage{algorithmic}

\newtheorem{thm}{Theorem}

\newtheorem{rem}{Remark}
\newtheorem{prop}{Proposition}

\usepackage{amsfonts,amsmath,amssymb,amstext}
\usepackage{color}

\def\diag{{\rm diag}}

\def\t0{{t_0}}

\def\H{{\mathcal H}}

\def\XX{{\mathcal X}}
\def\YY{{\mathcal Y}}
\def\ZZ{{\mathcal Z}}

\def\Nm{{\mathcal N}}
\def\R{{\mathbb R}}        
\def\Prob{{\bf Prob}}
\def\Proj{{\bf Proj}}

\def\im{{\rm im}}

\begin{document}
%
\title{HodgeRank with Information Maximization for \\ Crowdsourced Pairwise Ranking Aggregation} 


%
%

\author{Qianqian Xu$^{1,3}$, Jiechao Xiong$^{2,3}$, Xi Chen$^{4}$, Qingming Huang$^{5,6}$, Yuan Yao$^{7,3,\mbox{\Letter}}$\\
    $^1$ SKLOIS, Institute of Information Engineering, CAS, Beijing, China,  $^2$ Tencent AI Lab, Shenzhen, China\\
    $^3$ BICMR and School of Mathematical Sciences, Peking University, Beijing, China \\
    $^4$ Department of IOMS, Stern School of Business, New York University, USA\\
    $^5$ University of Chinese Academy of Sciences, Beijing, China, $^6$ IIP., ICT., CAS, Beijing, China\\
    $^{7,\mbox{\Letter}}$ Department of Mathematics, Hong Kong University of Science and Technology, Hong Kong\\
    xuqianqian@iie.ac.cn, jcxiong@tencent.com, xichen@nyu.edu, qmhuang@ucas.ac.cn, yuany@ust.hk $^{\mbox{\Letter}}$
}

\maketitle

\begin{abstract}

Recently, crowdsourcing has emerged as an effective paradigm for human-powered large scale problem solving in various domains. However, task requester usually has a limited amount of budget, thus it is desirable to have a policy to wisely allocate the budget to achieve better quality. In this paper, we study the principle of information maximization for active sampling strategies in the framework of HodgeRank, an approach based on Hodge Decomposition of pairwise ranking data with multiple workers. The principle exhibits two scenarios of active sampling: Fisher information maximization that leads to unsupervised sampling based on a sequential maximization of graph algebraic connectivity without considering labels; and Bayesian information maximization that selects samples with the largest information gain from prior to posterior, which gives a supervised sampling involving the labels collected. Experiments show that the proposed methods boost the sampling efficiency as compared to traditional sampling schemes and are thus valuable to practical crowdsourcing experiments.
\end{abstract}

\section{Introduction}\label{sec:introduction}

The emergence of online paid crowdsourcing platforms, like Amazon Mechanical Turk, presents us new possibilities to distribute tasks to human workers around the world,
on-demand and at scale. 
Recently, there arises a plethora of pairwise comparison
data in crowdsourcing experiments on Internet \cite{Tieyan11,xu2016false,chen2013pairwise,Yanwei14,XiChen}, where the comparisons can be modeled as oriented edges of an underlying graph. As online workers can come and complete tasks posted by
a company, and work for as long or as little as they wish, the data we collected are highly imbalanced where different alternatives might receive different number of comparisons,
and incomplete with large amount of missing values. To analyze
the imbalanced and incomplete data efficiently, the newly proposed Hodge theoretic approach \cite{Hodge} provides us a simple yet powerful tool.
%
%

HodgeRank, introduced by \cite{Hodge}, is an application of combinatorial Hodge theory to the preference or rank aggregation from pairwise comparison data. In an analog to Fourier decomposition in signal processing, Hodge decomposition of pairwise comparison data splits the aggregated global ranking and conflict of interests into different orthogonal components. It not only generalizes the classical Borda count in social choice theory to determine a global ranking from pairwise comparison data under various statistical models, but also
measures the conflicts of interests (i.e., inconsistency) in the pairwise comparison data. The inconsistency shows the validity of the ranking obtained
and can be further studied in terms of its geometric
scale, namely whether the inconsistency in the ranking
data arises locally or globally.

A fundamental problem in crowdsourcing ranking is the \emph{sampling} strategy, which is crucial to collect data efficiently. Typically, there are two ways to design sampling schemes: \emph{random sampling} and \emph{active sampling}. Random sampling is a basic type of sampling and the principle of random sampling is that every item has the same probability of being chosen at any stage during the sampling process.
The most important benefit of random sampling over active methods is its simplicity which allows flexibility and generality to diverse situations. However, this non-selective manner does not sufficiently use the information of past labeled pairs, thus potentially increases the costs in applications.
This motivates us to investigate efficient schemes for \emph{active sampling}.

In this paper, we present a principle of active sampling based on \emph{information maximization} in the framework of HodgeRank. Roughly speaking, Fisher's information maximization with HodgeRank leads to a scheme of unsupervised active sampling which does not depend on actual observed labels (i.e., a fixed sampling strategy before the data is observed). Since this sampling scheme does not need the feedback
from the worker, it is fast and efficient. Besides, it is insensitive to outliers.
On the other hand, a Bayesian information maximization equips us a supervised active sampling scheme that relies on the history of pairwise comparison data. By exploiting additional information in labels, supervised sampling often exhibits better performances than unsupervised active sampling and random sampling. However as the supervised sampling is sensitive to outliers, while reliability/quality of each worker is heterogeneous and unknown in advance, we find that unsupervised active sampling is sometimes more efficient than supervised sampling when the latter selects outlier samples at the initial stage.  Experimental results on both simulated examples and real-world data support the efficiency improvements of active sampling compared against passive random sampling.


 Our contributions in this work are threefold:

1. A new version of \emph{Hodge decomposition} of pairwise comparison data with multiple voters is presented. Within this framework, two schemes of information maximization, Fisher and Bayesian that lead to unsupervised and supervised sampling respectively, are systematically investigated.

2. Closed form update and a fast \emph{online algorithm} are derived for \emph{supervised sampling} with Bayesian information maximization for HodgeRank, which is shown faster and more accurate than the state-of-the-art method Crowd-BT \cite{chen2013pairwise}.

3. These schemes exhibit better sampling efficiency than random sampling as well as a better \emph{loop-free} control in clique complex of paired comparisons, thus reduce the possibility of causing voting chaos by harmonic ranking \cite{chaos} (i.e., the phenomenon that the inconsistency of preference data may lead to totally different aggregate orders using different methods).

\section{Hodge-theoretic approach to ranking}

\label{sec:framework0}
Before introducing our active sampling schemes, we will first propose a new version of Hodge decomposition of pairwise labels to ranking.

\subsection{From Borda count to HodgeRank}

In crowdsourced pairwise comparison experiments, let $V$ be the set of candidates and $|V| = n$. A voter (or worker)
$\alpha\in A$ provides his/her preference for a pair of
candidates $(i,j)\in V\times V$, $y^\alpha _{ij} : A\times V\times V\rightarrow \mathbb{R}$ such
that $y^\alpha_{ij}=-y_{ji}^\alpha$, where $y_{ij}^\alpha>0$ if $\alpha$ prefers $i$ to $j$
and $y_{ij}^{\alpha}\leq 0$ otherwise. The simplest setting is the binary choice, where
\begin{align}
y_{ij}^\alpha=\left\{\begin{array}{cc}
                                                  1 & \mathrm{if} \ \alpha \ \mathrm{prefers} \ i \ \mathrm{to} \ j, \\
                                                  -1 & \mathrm{otherwise}.
                                                \end{array}
                                                \right.
\label{eq:Y}
\end{align}

Such pairwise comparison data can be represented by a graph $G=(V,E)$, where $(i,j)\in E$ is an oriented edge when $i$ and $j$ are effectively compared by some voters. Associate each $(i,j)\in E$
a Euclidean space $\R^{|A_{ij}|}$ where $A_{ij}$ denotes the voters who compared $i$ and $j$. Now define $\YY:=\otimes_{(i,j)\in E} \R^{|A_{ij}|}$, 
a Euclidean space with standard basis $e_{ij}^\alpha$. In other words, for every pair of candidates, a vector space representing preferences of multiple voters or workers is attached to the corresponding graph edge, therefore $\YY$ can be viewed as a vector bundle or sheaf on the edge space $E$.

Statistical rank aggregation problem is to look for some
global rating score from such kind of pairwise comparison data. One of the well-known methods for this purpose is the Borda count in social choice theory \cite{Hodge}, in which the candidate that has the most pairwise comparisons in favour of it from all
voters will be ranked first, and so on. However, Borda count requires the data to be complete and balanced.
To adapt to new features in modern datasets, i.e. incomplete and imbalanced, the following least squares problem generalizes the classical Borda count to scenarios from complete to incomplete voting,

\begin{equation} \label{ls}
\min_{x} \|y - D_0 x\|_2^2
\end{equation}
where $x\in \XX:=\R^{|V|}$ is a global rating score, $D_0: \XX \rightarrow \YY$ is a finite difference (coboundary) operator defined by $(D_0 x)(\alpha,i,j) = x_i - x_j$. In other words, here we are looking for a universal rating model independent to $\alpha$, whose pairwise difference approximates the voter's data in least squares. We note that multiple models are possible if one hopes to group voters or pursue personalized ratings by extending the treatment in this paper.

Assume that $G$ is connected, then solutions of \eqref{ls} satisfy the following graph Laplacian equation which can be solved in nearly linear computational complexity \cite{spielman2004nearly,cohen2014solving}
\begin{equation} \label{laplacian}
D_0^T D_0 x = D_0^T y
\end{equation}
where $L=D_0^T D_0$ is the weighted graph Laplacian defined by $L(i,j) = -m_{ij}$ ($m_{ij}=|A_{ij}|$) for $i\neq j$ and $L(i,i) = \sum_{j : (i,j)\in E } m_{ij}$. The minimal norm least squares estimator is given by $\hat{x}=L^\dagger D_0^T y$ where $L^\dagger$ is the Moore-Penrose inverse of $L$.

\subsection{A new version of Hodge decomposition}
With the aid of combinatorial Hodge theory, the residue of \eqref{ls} can be further decomposed adaptive to the topology of clique complex $\chi_G=(V,E,T)$, where $T=\{(i,j,k): (i,j)\in E, (j, k)\in E, (k,i)\in E\}$ collects the oriented triangles (3-cliques) of $G$. To see this, define $\ZZ=\R^{|T|}$ and the triangular curl (trace) operator $D_1: \YY\to \ZZ$ by $(D_1y)(i,j,k)= \frac{1}{m_{ij}}\sum_{\alpha} y^{\alpha}_{ij} + \frac{1}{m_{jk}}\sum_{\alpha} y^{\alpha}_{jk} + \frac{1}{m_{ki}}\sum_{\alpha} y^{\alpha}_{ki} $. Plugging in the definition of $D_0$, it is easy to see $(D_1(D_0 x))(i,j,k) = (x_i - x_j) + (x_j - x_k) + (x_k - x_i) = 0$. In the following, we extend the existing HodgeRank methodology from simple graph with skew-symmetric preference to multiple digraphs with any preference, which potentially allows to model different users' behaviour. In particular, the existing Hodge decomposition \cite{Hodge} only considers the simple graph, which allows only one (oriented) edge between two nodes where pairwise comparisons are aggregated as a mean flow on the edge. However, in crowdsourcing applications, each pair is labeled by multiple workers. Therefore, there will be multiple inconsistent edges (edges in different directions) for each pair of nodes. Also the pairwise comparison data may not be skew-symmetric, for example home advantage of sports games. To meet this challenge, we need to extend existing theory to the following new version of Hodge decomposition theorem adapted to the multi-worker scenario.

\begin{thm}[Hodge Decomposition Theorem]\label{thm:hodge}
Consider chain map
\[ \XX \xrightarrow{D_0} \YY \xrightarrow{D_1} \ZZ \]
with the property $D_1\circ D_0=0$. Then for any $y\in \YY$, the following orthogonal decomposition holds
\begin{equation}\label{hodgedecomp}
y = b+u+ D_0 x + D_1^T z + w,
\end{equation}
\[ w\in \ker(D_0^T) \cap\ker(D_1), \]
where $b$ is the symmetric part of $y,~i.e.~b_{ij}^\alpha = b_{ji}^\alpha = (y_{ij}^\alpha+y_{ji}^\alpha)/2$, which captures the position bias of pairwise comparison on edge $(\alpha,i,j)$. The other four are skew-symmetric. $u$ is a universal kernel satisfying $\sum_\alpha u_{ij}^\alpha=0, \forall (i,j)\in E$ indicating all pairwise comparisons are completely in tie, $x$ is a global rating score, $z$ captures mean triangular cycles and $w$ is called harmonic ranking containing long cycles irreducible to triangular ones.
\end{thm}

The proof is provided in the supplementary materials. In fact all the components except $b$ and $D_0 x$ are of cyclic rankings, where the universal kernel $u$ as a complete tie is bi-cyclic for every edge $(i,j)\in E$. By adding 3-cliques or triangular faces to $G$, the clique complex $\chi_G$ thus enables us to separate the triangle cycles $D_1^T z$ from the cyclic rankings. Similarly one can define dimension-2 faces of more nodes, such as quadrangular faces etc., to form a \emph{cell complex} to separate high order cycles via Hodge decomposition. Here we choose clique complex $\chi_G$ for simplicity.   
The remaining harmonic ranking $w$ is generically some long cycle involving all the candidates in comparison, therefore it is the source of voting or ranking chaos \cite{chaos} (a.k.a. fixed tournament issue in computer science), i.e., any candidate $i$ can be the final winner by removing some pairwise comparisons containing the opponent $j$ who beats $i$ in such comparisons. Fortunately harmonic ranking can be avoided by controlling the topology of underlying simplicial complex $\chi_G$; in fact Hodge theory tells us that harmonic ranking will vanish if the clique complex $\chi_G$ (or cell complex in general) is loop-free, i.e., its first Betti number being zero. In this case, the harmonic ranking component will be decomposed into local cycles such as triangular cycles. Therefore in applications it is desired to have the simplicial complex $\chi_G$ loop free, which is studied later in this paper with active sampling schemes. For this celebrated decomposition, the approach above is often called \emph{HodgeRank} in literature.

When the preference data $y$ is skew-symmetric, the bias term $b$ vanishes, there only exists a global rating score and cyclic rankings. Cyclic rankings part mainly consists of noise and outliers, where outliers have much larger magnitudes than normal noise. So a sparse approximation of the cyclic rankings for pairwise comparison data can be used to detect outliers. In a mathematical way, suppose $\Proj$ is the projection operator to the cyclic ranking space, then $\Proj(\gamma)$ with a sparse outlier vector $\gamma$ is desired to approximate $\Proj(y)$. One popular method is LASSO as following:
\[\min_{\gamma} \|\Proj(y) - \Proj(\gamma)\|_2^2 + \lambda\|\gamma\|_1\]

Further more, the term $b$ models the user-position bias in the preference. It means on the edge $(\alpha,i,j)$ and $(\alpha,j,i)$, there is a bias caused by various reasons, such as which one is on the offensive. While in most crowdsourcing problems, we believe there should not have a such term unless the worker is careless. So this term can be used to model the workers' behavior. In formulation, we can add an intercept term into \eqref{ls}:
\begin{equation}
\min_{x} \|y - b - D_0 x\|_2^2
\end{equation}
where $b$ is a piecewise constant intercept depending on worker $\alpha$ only: $b_{ij}^\alpha = constant_\alpha, \forall i,j$. Such an intercept term can be seen as a mean effect of the position bias for each worker. The bigger its magnitude is, the more careless the worker is. Generally, this term can be any piecewise constant vector which models different group effect of bias. This potentially allows to model different workers' behavior.

\subsection{Statistical models under HodgeRank}
HodgeRank provides a unified framework to incorporate various statistical models, such as Uniform model, Thurstone-Mosteller model, Bradley-Terry model, and especially Mosteller's Angular Transform model which is essentially the only model having the asymptotic variance stabilization property. These are all generalized linear models for binary voting. In fact, generalized linear models assume that the probability of pairwise preference is fully decided by a linear function as follows\begin{equation}
\pi_{ij} = \Prob\{\text{$i\succ j$} \} = \Phi(x^\ast_i - x^\ast_j), \ x^\ast\in \XX
\end{equation}
where $\Phi:\R \to [0,1]$ can be chosen as any symmetric cumulated distributed function. In a reverse direction, if an empirical preference probability $\hat{\pi}_{ij}$ is observed in experiments, one can map $\hat{\pi}$ to a skew-symmetric pairwise comparison data by the inverse of $\Phi$, $\hat{y}_{ij} = \Phi^{-1} ( \hat{\pi}_{ij} )$. Then solving the HodgeRank problem \eqref{ls} is actually solving the weighted least squares problem for this generalized linear model. Different choices of $\Phi$ lead to different generalized linear models, e.g. $\Phi(t)=e^t/(1+e^t)$ gives Bradley-Terry model and $\Phi(t)=(\sin(t)+1)/2$ gives Mosteller's Angular Transform model.

\section{Information Maximization for Sampling in HodgeRank} \label{sec:framework}

Our principle for active sampling is \emph{information maximization}. Depending on the scenarios in application, the definition of \emph{information} varies. There are often two  ways to design active sampling strategies depending on available information: (1) unsupervised active sampling without considering the actual labels collected, where we use Fisher information  to maximize algebraic connectivity in graph theory; (2) supervised active sampling with label information, where we exploit a Bayesian approach to maximize expected information.
In the following, we will first introduce the unsupervised active sampling, followed by the
supervised active sampling.  After that, an online algorithm of supervised active
sampling will be detailed. Finally, we discuss the online tracking of topology evolutions of the sampling schemes.

\subsection{Fisher information maximization: unsupervised sampling}

In case that the cyclic rankings in \eqref{hodgedecomp} are caused by Gaussian noise, i.e. $u + D_1^T z + w=\epsilon$ where $\epsilon\sim \Nm(0,\Sigma_\epsilon)$, the least squares problem \eqref{ls} is equivalent to the following Maximum Likelihood problem:
%
\[\max_{x}\frac{(2\pi)^{-m/2}}{\det(\Sigma_\epsilon)} \exp\left(-\frac{1}{2}(y - D_0 x)^T\Sigma_{\epsilon}^{-1}(y - D_0 x)\right),\]
where $\Sigma_{\epsilon}$ is the covariance matrix of the noise, $m = \sum_{(i,j)\in E} m_{ij}$. In applications without a priori knowledge about noise, we often assume the noise is independent and has unknown but fixed variance $\sigma_\epsilon^2$, i.e. $\Sigma_{\epsilon} = \sigma_\epsilon I_m$. So HodgeRank here is equivalent to solve the Fisher's Maximum Likelihood with Gaussian noise. Now we are ready to present a sampling strategy based on Fisher information maximization principle.

{\bf Fisher Information Maximization}:
The log-likelihood is
\[l(x) = -m\log (\sqrt{2\pi}\sigma_\epsilon) -\frac{1}{2}(y - D_0 x)^T\Sigma_{\epsilon}^{-1}(y - D_0 x).\] So the Fisher Information is given as
\begin{equation}
I := -E\frac{\partial^2 l}{\partial x^2}=   D_0^T \Sigma_{\epsilon}^{-1} D_0 = L/\sigma_\epsilon^2.
\end{equation}
where $L =D_0^T D_0$ is  the weighted graph Laplacian.

Given a sequence of samples $\{\alpha_t, i_t,j_t\}_{t\in N}$ (edges), the graph Laplacian can be defined recursively as $L_{t} = L_{t-1}+d_{t}^Td_{t}$, where $d_{t}:\XX \to \YY$ is defined by $(d_t x)(\alpha_t,i_t,j_t)= x_{i_t}-x_{j_t}$ and $0$ otherwise. Our purpose is to maximize the Fisher information given history via
\begin{equation} \label{fisher}
\max_{(\alpha_t,i_{t},j_{t})} f(L_{t})
\end{equation}
where $f:S_+^n \rightarrow R $ is a concave function w.r.t the weights on edges. Since it is desired that the result does not depend on the index $V$, $f$ has to be permutation invariant. A stronger requirement is orthogonal invariant $f(L)=f(O^T L O)$ for any orthogonal matrix $O$, which implies that $f(L) = g(\lambda_2,\dots,\lambda_n)$, $0=\lambda_1 \le \lambda_2\le\dots\le\lambda_n$ are the eigenvalues of $L$ \cite{Pablo12}. Note that it does not involve sampling labels and is thus an unsupervised active sampling scheme.

\begin{figure}[t!]
\begin{center}

\includegraphics[width=0.3\textwidth]{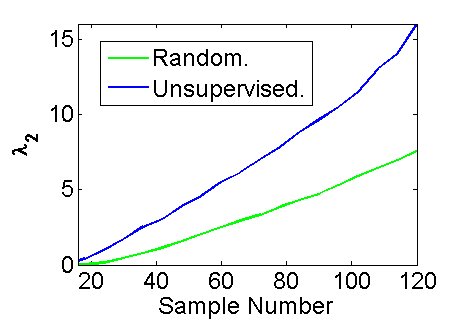}

\caption{Fiedler value comparison of unsupervised active sampling vs. random sampling.}
\label{icml-simulatedvalid}
\end{center}
\end{figure}

Among various choices of $f$, a popular one is $f(L_t) = \lambda_2(L_t)$, where $\lambda_2(L_t)$ is the smallest nonzero eigenvalue (a.k.a. algebraic connectivity or Fiedler value) of $L_t$, which corresponds to ``E-optimal" in experimental design \cite{osting2014optimal2}. Despite that \eqref{fisher} is a convex optimization problem with respect to real-valued graph weights, the optimization over integral weights is still NP-hard and a greedy algorithm \cite{GhoshBoyd06} can be used as a first-order approximation
\begin{eqnarray*}
\max \lambda_2(L_t)  &\approx &\max [\lambda_2(L_{t-1}) + \|d_t v_2(L_{t-1}) \|^2] \\
&= & \lambda_2(L_{t-1})+\max (v_2(i_t) - v_2(j_t))^2,
\end{eqnarray*}
where $v_2$ is the second nonzero eigenvector or Fiedler vector of $L_{t-1}$.
Figure \ref{icml-simulatedvalid} shows Fiedler value plots of two sampling schemes, where unsupervised active sampling above effectively raises the Fiedler value curve than random sampling.

While the unsupervised sampling process only depends on $L_t$, label information is collected for the computation of HodgeRank global ranking estimator
$\hat{x}^t = L_t^{\dagger} (D_{0}^t)^T y^t$, where $D_{0}^t= D^{t-1}_0 + d_t$ and $y^t = y^{t-1}+y^{\alpha_t}_{i_t j_t} e^{\alpha_t}_{i_t j_t}$.


\begin{algorithm}
	\caption{Unsupervised active sampling algorithm.}
	\label{alg-Unsupervised}
	\KwIn{An initial graph Laplacian $L_0$ defined on the graph of $n$ nodes.}
	\For{$t=1,\dots,T$}
	{
		Compute the second eigenvector $v_2$ of $L_{t-1}$.\;
		Select the pair $(i_t, j_t)$ which maximizes $(v_2(i_t) - v_2(j_t))^2$.\;
		Draw a sample on the edge $(i_t, j_t)$ with voter $\alpha_t$.\;
		Update graph Laplacian $L_t$.\;
	}
	\KwOut{Sampling sequence $\{\alpha_t, i_t,j_t\}_{t\in N}$.}
\end{algorithm}

\subsection{Bayesian information maximization: supervised sampling}

Since the least squares problem \eqref{ls} is invariant under the shift of $x$, a small amount of regularization is always preferred.
Therefore in practice \eqref{ls} is understood as the minimal norm least squares solution, or the ridge regularization,
\begin{equation}\label{ridge}
\min_{x} \|y - D_0 x\|_2^2 + \gamma \|x\|_2^2.
\end{equation}
Regularization on $x$ means a prior distribution assumption on $x$. So \eqref{ridge} is equivalent to
\begin{equation}\label{eq:l2hodge}
\max_{x} \exp\left(-\frac{\|y - D_0 x\|_2^2}{2\sigma_\epsilon^2}-\frac{\|x\|_2^2}{2\sigma_x^2}\right),
\end{equation}
when $\sigma_\epsilon^2/\sigma_x^2 = \gamma$. 
So regularized HodgeRank is equivalent to the Maximum A Posterior (MAP) estimator when both the likelihood and prior are Gaussian distributions.


With such a Bayesian perspective, a natural scheme for active sampling is based on the maximization of expected information gain (EIG) or Kullback-Leibler divergence from prior to posterior. In each step, the most informative triplet (object $i$, object $j$, annotator $\alpha$) is added based on the largest KL-divergence between posterior and prior.
The maximization of EIG  has been a popular criterion in active sampling  \cite{Settles:09} and  applied to some specific pairwise comparison models (e.g. \cite{chen2013pairwise} applied EIG to Bradley-Terry model with Gaussian prior and \cite{Thomas:12} to Thurstone-Mosteller model).
Combining the EIG criterion with the $\ell_2$-regularized HodgeRank formulation in \eqref{eq:l2hodge}, we obtain a simple closed form update for the posterior for general models, which leads to a fast online algorithm.

{\bf Bayesian information maximization}: Let $P^{t}(x|y^{t})$ be the posterior of $x$ given data $y^{t}$. So given present data $y^{t}$, we choose a new pair to maximize the expected information gain (EIG) of a new pair $(i,j)$:
\begin{equation}\label{eq:i_j_EIG}
(i^*,j^*) = \arg \max_{(i,j)} EIG_{(i,j)}
\end{equation}
where
\begin{equation} \label{EIG}
EIG_{(i,j)} := E_{y^{t+1}_{ij}|y^{t}} KL(P^{t+1}| P^{t})
\end{equation}
and the KL-divergence
\[KL(P^{t+1}| P^{t}) := \int P^{t+1}(x|y^{t+1}) \ln\frac{P^{t+1}(x|y^{t+1})}{P^{t}(x|y^{t})}dx\]

Once an optimal pair $(i^*,j^*)$ is determined from \eqref{eq:i_j_EIG}, we assign this pair to a random voter $\alpha \in A$ and then collect the corresponding label for the next update.

In the $l_2$-regularized HodgeRank setting, such a optimization problem in \eqref{eq:i_j_EIG} can be greatly simplified.
\begin{prop}\label{prop1}
When both the likelihood and prior are Gaussian distributions, then posterior $P^t(x|y^{t})$ is also Gaussian.
\[  x|y^{t} \sim N(\mu^t,\sigma_\epsilon^2\Sigma^t )\]
\[\mu^t= (L_t + \gamma I)^{-1}(D_0^t)^Ty^t, \Sigma^t= (L_t + \gamma I)^{-1}. \]
Thus
\begin{eqnarray}\label{KL1}
&&2KL(P^{t+1}| P^{t}) \\ \nonumber
&=&\ \frac{1}{\sigma_\epsilon^2}(\mu^t-\mu^{t+1})^T(L_t+\gamma I) (\mu^t-\mu^{t+1})  - n\\ \nonumber
&&\ + tr((L_t+\gamma I)(L_{t+1}+\gamma I)^{-1}) \\
&&\ + \ln\frac{\det(L_{t+1}+\gamma I)}{\det(L_t+\gamma I)}
\end{eqnarray}

and the posterior $y_{ij}^{t+1}|y^{t} \sim N(a,b)$ with $a = \mu^t_i-\mu^t_j, b = (\Sigma_{ii}^t+\Sigma_{jj}^t-2\Sigma_{ij}^t+1)\sigma_\epsilon^2.$

\end{prop}

\begin{rem}
Note the first term of $KL(P^{t+1}| P^{t})$ is $l_2$ distance of gradient flow between $\mu^t$ and $\mu^{t+1}$ if $\gamma =0 $. The unknown parameter $\sigma_\epsilon$ needs a roughly estimation. In binary comparison data, $\sigma_\epsilon = 1$ is good enough. Given the history $D_0^t, y^t$ and the new edge $(i,j)$, $\mu^{t+1}$ is only a function of $y^{t+1}_{ij}$, so does $KL(P^{t+1}| P^{t})$.
\end{rem}

Generally, the posterior of $y_{ij}^{t+1}$
\[p(y^{t+1}_{ij}|y^{t}) = \int p(y_{ij}^{t+1}|x)P^t(x|y^{t})dx\]
can be approximated by $p(y_{ij}^{t+1}|\hat{x}^t)$, where $\hat{x}^t$ is the HodgeRank estimator $\mu^t$. In practice, we receive binary comparison data $y^\alpha_{ij}\in \{\pm 1\}$, hence we can adopt generalized additive models $\hat{\pi}(y^\alpha_{ij}=1) = \Phi(\hat{x}_i -\hat{x}_j)$ to compute it explicitly.

Such a Bayesian information maximization approach relies on actual labels collected in history, as sampling process depends on $y^t$ through $\mu^t$. Hence it is a supervised active sampling scheme, in contrast to the previous one.
%
%

\subsection{Online supervised active sampling algorithm}

To update the posterior parameters efficiently, we would like to introduce an accelerating method using Sherman-Morrison-Woodbury formula \cite{bartlett1951inverse}.  In active sampling scheme, the Bayesian information maximization approach needs to compute EIG for ${n \choose 2}$ times to choose one pair. And each EIG consists of the computation of inverting an $n\times n$ matrix, which costs $O(n^3)$ and is especially expensive for large scale data. But notice that $L_{t+1}$ and $L_{t}$ only differs by a symmetric rank-1 matrix, Sherman-Morrison-Woodbury formula can be applied to greatly accelerate the sampling procedure.

Denote $L_{t,\gamma}=L_t + \gamma I$, so $L_{t+1,\gamma} = L_{t,\gamma}+d_{t+1}^Td_{t+1}$, then Sherman-Morrison-Woodbury formula can be rewritten as follows:
\begin{equation} \label{SMW}
L_{t+1,\gamma}^{-1} = L_{t,\gamma}^{-1} - \frac{L_{t,\gamma}^{-1}d_{t+1}^Td_{t+1}L_{t,\gamma}^{-1}}{1 + d_{t+1}L_{t,\gamma}^{-1}d_{t+1}^T}
\end{equation}

\begin{prop}\label{prop2}
Using the Sherman-Morrison-Woodbury formula, Eq (\ref{KL1}) can be further simplified as
\begin{eqnarray} \label{KL-explicit}
&&KL(P^{t+1}|P^t)\\
&=&\frac{1}{2}[\frac{1}{\sigma_\epsilon^2}(\frac{y_{ij}^{t+1}-d_{t+1}\mu^t}{1+C})^2 C + \ln( 1 - C) - \frac{C}{1 + C}] \nonumber
\end{eqnarray}
where $C = d_{t+1}L_{t,\gamma}^{-1}d_{t+1}^T$ and
\begin{eqnarray}\label{update-score}
\mu^{t+1}
&=&\mu^{t} + \frac{y_{ij}^{t+1}-d_{t+1}\mu^t}{1+C} L_{t,\gamma}^{-1}d_{t+1}^T.
 \end{eqnarray}
\end{prop}



Now for each pair of nodes $(i,j)$, we only need to compute $d_{t+1}L_{t,\gamma}^{-1}d_{t+1}^T$ and $d_{t+1}\mu^t$. Since $d_{t+1}$ has the form of $e_i-e_j$, so it only costs $O(1)$ which is much cheaper than the original $O(n^3)$. The explicit formula of KL-divergence (\ref{KL-explicit}) makes the computation of EIG easy to vectorize, especially useful in MATLAB. Also note that if we can store the matrix $L_{t,\gamma}^{-1}$, (\ref{SMW}) provides the formula to update $L_{t,\gamma}^{-1}$ and (\ref{update-score}) provides the update of score function $\mu^t$. Combining these two posterior update rules,  the entire
online active algorithm is presented in Algorithm \ref{alg-online}.


%

\begin{algorithm}
	\caption{Online supervised active sampling algorithm for binary comparison data.}
	\label{alg-online}
	\KwIn{Prior distribution parameters $\gamma, \mu^0, L_{0,\gamma}^{-1}$.}
	\For{$t=0,1,\dots,T-1$}
	{
		For each pair $(i,j)$, compute the expected information gain in Eq. (\ref{EIG}) and Eq. (\ref{KL-explicit}) using $\sigma_{\epsilon}=1$\;
		Select the pair $(i^\ast, j^\ast)$ which has maximal EIG.\;
		Draw a sample on the edge $(i^\ast, j^\ast)$ from a randomly chosen voter $\alpha_t$ and observe  the next label $y_{i^\ast j^\ast}^{t+1}$.\;
		Update posterior parameters according to (\ref{SMW}) and (\ref{update-score}).\;
	}
	\KwOut{Ranking score function $\mu^T$.}
\end{algorithm}

%
%

\subsection[Online tracking of topology evolution]
{Online tracking of topology evolution}
In HodgeRank, two topological properties of clique complex $\chi_G$ have to be considered which are obstructions for obtaining global ranking and harmonic ranking.  First of all, a global ranking score can be obtained, up to a translation, only if the graph $G$ is connected, so one needs to check the number of connected components as the zero-th Betti number $\beta_0$. Even more importantly, the voting chaos indicated by harmonic ranking $w$ in \eqref{hodgedecomp} vanishes if the clique complex is loop-free, so it is necessary to check the number of loops as the first Betti number $\beta_1$. Given a stream of paired comparisons, persistent homology \cite{EdeLetZom02,Carlsson09} is in fact an online algorithm to check topology evolution when simplices (e.g. nodes, edges, and triangles) enter in a sequential way such that the subset inclusion order is respected.
Here we just discuss in brief the application of persistent homology to
monitor the number of connected components ($\beta_0$) and loops
($\beta_1$) in three different sampling settings.

\begin{figure}[t!]
\begin{center}
\subfigure [Unsupervised.]{
\includegraphics[width=0.14\textwidth]{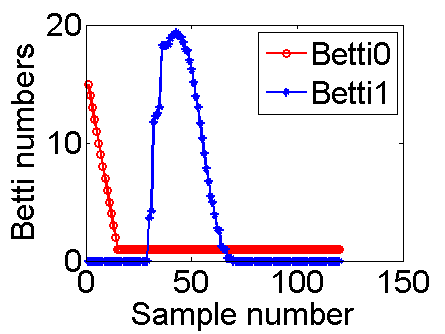}}
\subfigure [Supervised.]{
\includegraphics[width=0.14\textwidth]{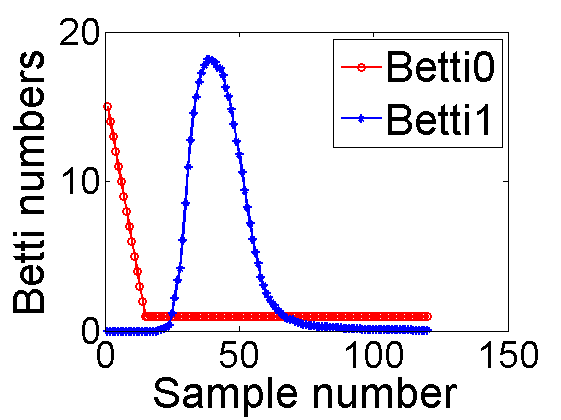}}
\subfigure [Random.]{
\includegraphics[width=0.14\textwidth]{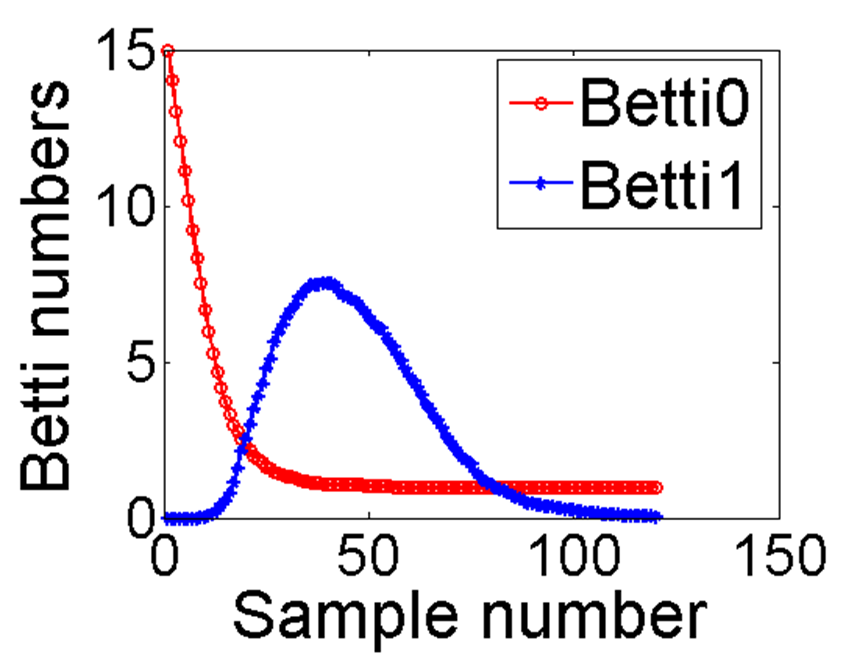}}

\caption{Average Betti numbers for three sampling schemes.}
\label{persistent_homo}
\end{center}
\end{figure}

Assume that the nodes come in a
certain order (e.g., production time, or all created in the same
time), after that pairs of edges are presented to us
one by one guided by the corresponding sampling scheme. A triangle $\{i,j,k\}$ is created whenever
all the three associated edges appeared. Persistent homology may
return the evolution of the number of connected components
($\beta_0$) and the number of independent loops ($\beta_1$) at each
time when a new node/edge/triangle is born. The expected
$\beta_0$ and $\beta_1$ (with 100 graphs) computed by Javaplex \cite{jplex} for $n = 16$ of three sampling schemes are plotted in
Figure \ref{persistent_homo}. It is easy to see that both unsupervised \& supervised active sampling schemes narrow the nonzero region of $\beta_1$, which indicates that these two active sampling schemes both enlarge the loop-free regions thus reduce the chance of harmonic ranking or voting chaos.

\section{Experiments} \label{sec:experiments}

In this section, we study examples with both simulated and real-world data to illustrate the validity of the proposed two schemes of active sampling.

 \begin{figure}[t!]
\begin{center}
 \centering
\includegraphics[width=0.3\textwidth]{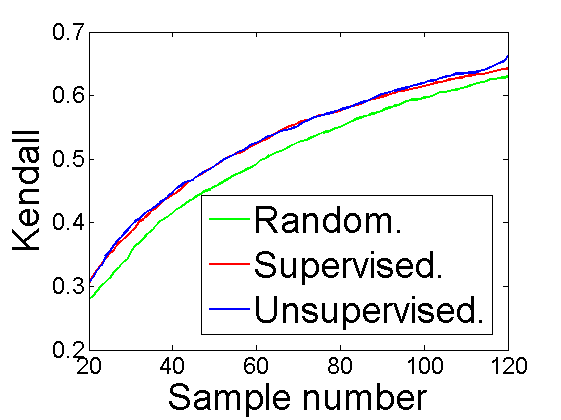}

\caption{The mean Kendall's $\tau$ between ground-truth and HodgeRank estimator for three sampling schemes.}
\label{three_simulated}
\end{center}
\end{figure}

\subsection{Simulated data}

In this experiment, we use simulated data to illustrate the performance differences among unsupervised \& supervised active sampling, and random sampling.
 We first randomly create a global ranking score $x$ as the ground-truth, uniformly distributed on [0,1] for $n$ candidates. Then we sample pairs from this graph using these three sampling schemes. The pairwise comparisons are generated by uniform model, i.e. $y^{\alpha}_{ij} = 1$ with probability $(x_i-x_j+1)/2$, $y^{\alpha}_{ij}=-1$ otherwise. Averagely, there are $30\% - 35\%$ comparisons are in the wrong direction, $(x_i-x_j)y^\alpha_{ij} < 0$. The experiments are repeated 1000 times and ensemble statistics for the HodgeRank estimator are recorded.

\begin{itemize} [leftmargin=0.01em,noitemsep,topsep=0pt]

\item \textbf{Kendall's $\tau$ comparison.} First, we adopt the
Kendall rank correlation ($\tau$) coefficient \cite{KendallYao} to measure the rank correlation
between ground-truth and HodgeRank estimator of these three sampling schemes. Figure \ref{three_simulated} shows the mean Kendall's $\tau$ associated with these three sampling schemes for $n = 16$ (chosen to be consistent with the first two real-world datasets considered later). The $x$-axes of the graphs are the number of samples added, taken to be greater than $\frac{\log n}{n}$ percentage so that the random graph is connected with high probability. From these experimental results, we observe that both active sampling schemes, with a similar performance, show better
efficiency than random sampling with higher Kendall's $\tau$.

%

\item \textbf{Computational cost.}  Table \ref{tableer} shows the computational complexity achieved by online/offline algorithms of supervised active sampling and unsupervised active sampling.
The total number of edges added is ${n \choose 2}$ and the value in this table represents the average time (s) needed of 100 runs for different \emph{n}. All computation is done using MATLAB R2014a, on a Mac Pro desktop PC, with 2.8 GHz Intel Core i7-4558u, and 16 GB memory.
It is easy to see that online supervised algorithm is faster than unsupervised active sampling. Besides, it can achieve up to hundreds of times faster than offline supervised algorithm, with exactly the same performances. And more importantly, as $n$ increases, such a benefit is increasing, which further implies its advantage in dealing with large-scale data.

\begin{figure}[t!]
\begin{center}
\subfigure [T=K=120]{
\includegraphics[width=0.31\linewidth]{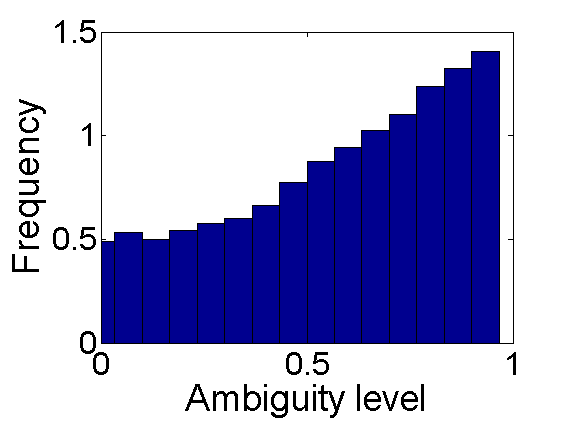}}
\subfigure [T=5K=600]{
\includegraphics[width=0.31\linewidth]{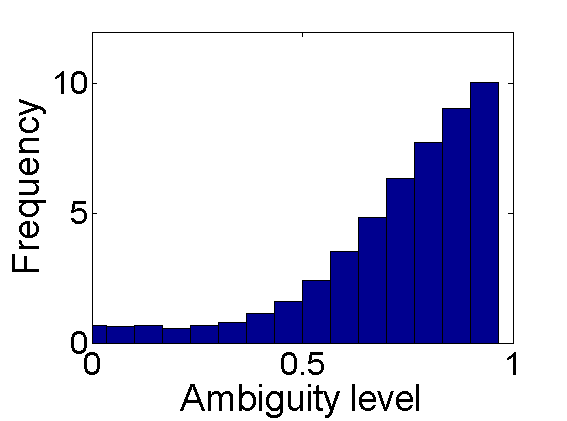}}
\subfigure [T=10K=1200]{
\includegraphics[width=0.31\linewidth]{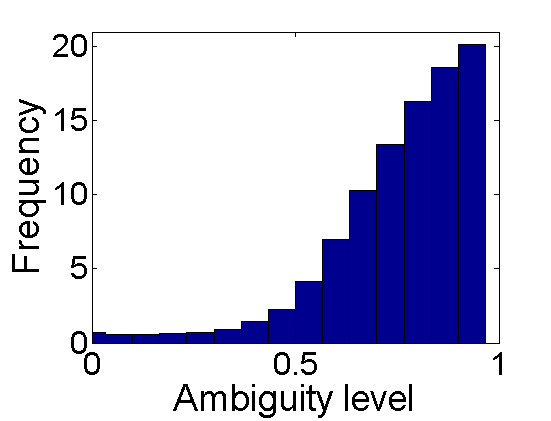}}

\caption{Sampling counts for pairs with different levels of ambiguity in supervised active sampling.}
\label{sample_frequency}
\end{center}
\end{figure}

\begin{table}\caption{\label{tableer} Computational complexity (s) comparison on simulated data.}

\centering
\newsavebox{\tablebox}
\begin{lrbox}{\tablebox}
 \begin{tabular}{ccccccc}
  \hline  $\emph{\textbf{n}}$   &\textbf{16} &\textbf{20} &\textbf{24} &\textbf{28} &\textbf{32} &\textbf{100}\\
  \hline
 \hline   \textbf{Offline Sup.}  &25.22 &81.65 &225.54 &691.29 &1718.34 &$>$7200  \\

  \hline  \textbf{Online Sup.}  &0.10 &0.17 &0.26 &0.38 &0.50 &15.93  \\

  \hline  \textbf{Unsup.}  &0.75 &1.14 &4.27 &6.73 &9.65 &310.58  \\
 \hline
 \end {tabular}
 \end{lrbox}
\scalebox{0.75}{\usebox{\tablebox}}
\end{table}

\item \textbf{Budget Level.} Next, we would like to investigate how the total budget is allocated among pairs with different levels of ambiguity in supervised active sampling scheme. In particular, we first randomly create a global ranking score as the ground truth,
uniformly distributed on [0, 1] for $n$ candidates, corresponding to $K$ = $n \choose 2$ pairs with different ambiguity levels (i.e., 1-abs (ground-truth score differences). In this experiment, $n = 16$, we vary the total budget $T = K, 5K, 10K$, and report
the number of times that each pair is sampled on average over 100 runs. The results are presented in Figure \ref{sample_frequency}.
It is easy to see that more ambiguous pairs with \emph{Ambiguity Level} close to 1 in general receive more labels than those simple pairs close to 0. This is consistent with practical applications, in which we should not spend too much budget on those easy pairs, since they can be decided based on the common knowledge and majority voting, excessive efforts will not bring much
additional information.

\end{itemize}

\begin{figure}[t!]
\begin{center}
 \centering
\includegraphics[width=\linewidth]{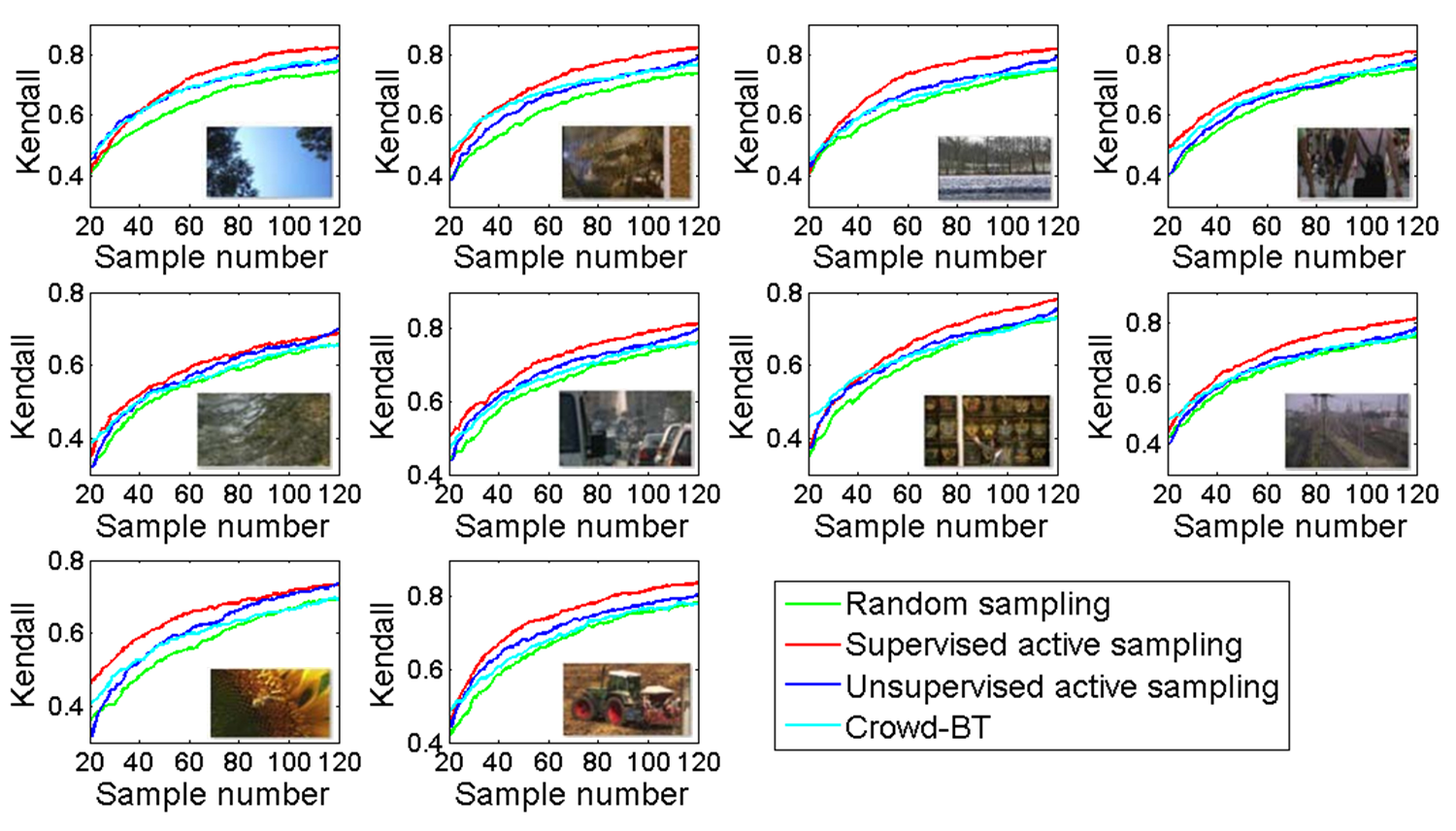}
\caption{Experimental results of four sampling schemes for 10 reference videos in LIVE database.}
\label{VQA}
\end{center}
\end{figure}

\subsection{Real-world data}


The first example gives a comparison
of these three sampling schemes on a complete \& balanced video quality assessment (VQA) dataset \cite{mm11}.
It contains 38,400 paired comparisons of the LIVE dataset \cite{LIVE} from 209 random observers.
As there is no ground-truth scores available, results obtained from all the paired comparisons are treated as the ground-truth. To ensure the statistical stability,
for each of the 10 reference videos, we sample using each of the three methods for 100 times. For comparison,
we also conduct experiments with the state-of-the-art method Crowd-BT \cite{chen2013pairwise}.
Figure \ref{VQA} shows the results and these different reference videos exhibit similar observations.
Consistent with the simulated data, the proposed unsupervised/supervised active sampling performs better
than random sampling scheme in the prediction of global ranking scores, and
the performance of supervised active sampling is slightly better than unsupervised active sampling with higher kendall's $\tau$. Moreover, our supervised active sampling consistently
manages to improve the kendall's $\tau$ of Crowd-BT by roughly 5\%.

\begin{figure}[t!]
\begin{center}
 \centering
\includegraphics[width=\linewidth]{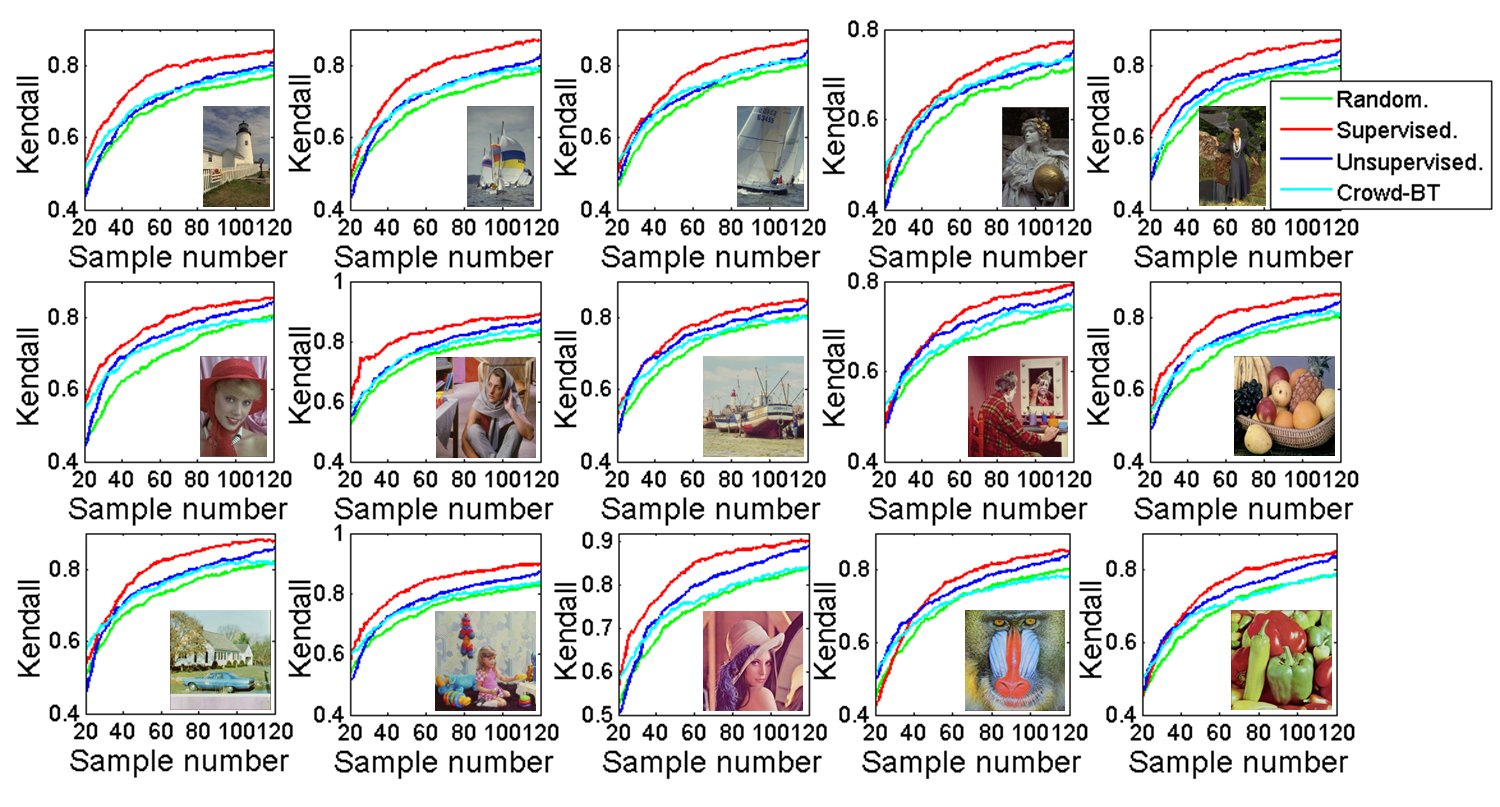}

\caption{Experimental results of four sampling schemes for 15 reference images in LIVE and IVC databases.}
\label{IQA}
\end{center}
\end{figure}

The second example shows the sampling results on an imbalanced dataset for image quality assessment (IQA), which contains
43,266 paired comparisons of 15 reference images \cite{LIVE}\cite{IVC} from 328 random observers on Internet.
As this dataset is relatively large and edges occurred on each paired comparison graph  with 16 nodes are dense, all the 15 graphs are also complete graph,
though possibly imbalanced. Figure \ref{IQA} shows mean Kendall's $\tau$ of 100 runs, and similarly for all reference images active sampling schemes show better performance than random sampling. Besides, our proposed supervised active sampling also performs better than Crowd-BT.

In the third example, we test our method to the task of ranking documents
by their reading difficulty. This dataset \cite{chen2013pairwise} is composed of 491 documents. Using the \textbf{CrowdFlower} crowdsourcing platform, 624 distinct annotators from the United States and Canada provide us a total of 12,728 pairwise comparisons. For better visualization, we only present the mean Kendall's $\tau$ of 100 runs
for the first 4,000 pairs in Figure \ref{document}. As captured
in the figure, the proposed supervised active strategy significantly outperforms
the random strategy. We also compare our method with
Crowd-BT and it is easy to see that our method also improves
over the Crowd-BT method's performance.

\begin{itemize} [leftmargin=0.01em,noitemsep,topsep=0pt]
\item \textbf{Running cost.}
More importantly, our method is much faster than Crowd-BT by orders of magnitude due to closed-form posterior in Proposition 1 and fast online computation in Proposition 2. Table \ref{tableer2} shows the comparable computational cost of these two methods using the same settings with Table \ref{tableer}. It is easy to see that On VQA dataset, for a reference video, 100 runs of Crowd-BT take about 10 minutes on average; while our online supervised algorithm takes only 18 seconds, which is 33 times faster. Besides, our method can achieve nearly 40 times speed-up on IQA dataset and 35 times faster on reading level dataset. In a word, the main advantages of our
method lies in its computational efficiency and the ability
to handle streaming data.

\begin{figure}[t!]
\begin{center}
 \centering
\includegraphics[width=0.3\textwidth]{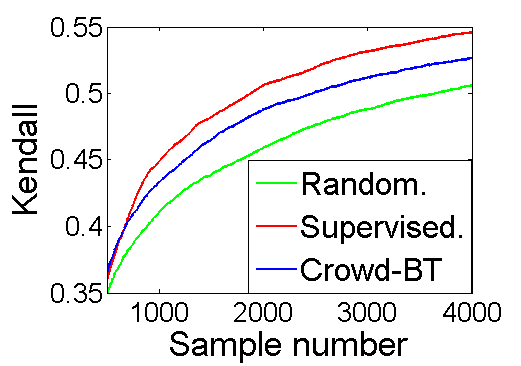}

\caption{Experimental results of three sampling schemes on reading level dataset.}
\label{document}
\end{center}
\end{figure}

\item \textbf{ Parameter tuning.} A crucial question here is how to choose $\gamma$ in supervised active sampling experiments. In practice, for dense graph, we find that $\gamma$ makes little difference in the experimental results, a smaller $\gamma$, say $0.01$, or even $1e^{-5}$ is sufficient. However, for sparse graph such as reading level dataset, a bigger $\gamma$ (i.e., $\gamma =1 $) may produce better performance.

\end{itemize}

%

\begin{table}\caption{\label{tableer2} Average running cost (s) of 100 runs on three real-world datasets.}

\centering

\begin{lrbox}{\tablebox}
 \begin{tabular}{ccc}
  \hline  $\emph{\textbf{Method}}$   &\textbf{Our supervised method} &\textbf{Crowd-BT}\\
  \hline
 \hline   \textbf{VQA dataset}  &18 &600  \\

  \hline  \textbf{IQA dataset}  &12 &480  \\

  \hline  \textbf{Reading level dataset}  &120 &4200  \\
 \hline
 \end {tabular}
 \end{lrbox}
\scalebox{0.75}{\usebox{\tablebox}}
\end{table}


\section{Conclusions} \label{sec:conclusions}
In this paper, we proposed a new Hodge decomposition of pairwise comparison data with multiple voters and analyzed two active sampling schemes in this framework. In particular, we showed that: 1) for unsupervised active sampling without considering
the actual labels, we can use Fisher information to maximize
algebraic connectivity in graph theory; 2) for supervised
active sampling with label information, we can exploit a
Bayesian approach to maximize expected information gain from prior to posterior. The unsupervised sampling involves the computation of a particular eigenvector of graph Laplacians, the Fiedler vector, which can be precomputed \emph{a priori}; while the supervised sampling benefits from a fast online algorithm using the Sherman-Morrison-Woodbury formula for matrix inverses, which however depends on the label history. Both schemes enable us a more efficient budget control than passive random sampling, tested with both simulated and real-world data, hence provide a helpful tool for researchers who exploit crowdsourced
pairwise comparison data.

\section{Acknowledgments}

The research of Qianqian Xu was supported by National Key Research and Development Plan (No.2016YFB0800403), National Natural Science Foundation of China (No.U1636214, 61422213, 61672514, 61390514, 61572042), CCF-Tencent Open Research Fund. The research of Xi Chen was supported in part by Google Faculty Research Award and Adobe Data Science Research Award. The research of Qingming Huang was supported in part by National Natural Science Foundation of China: 61332016, U1636214, 61650202 and 61620106009, in part by Key Research Program of Frontier Sciences, CAS: QYZDJ-SSW-SYS013. The research of Yuan Yao was supported in part by Hong Kong Research Grant Council (HKRGC) grant 16303817, National Basic Research Program of
China (No. 2015CB85600, 2012CB825501), National Natural Science Foundation of China (No. 61370004, 11421110001), as well as awards
from Tencent AI Lab, Si Family Foundation, Baidu BDI, and Microsoft Research-Asia. 

\normalsize
\bibliographystyle{aaai}

\bibliography{sigproc}

\section*{Supplementary Materials}

\appendix

\subsection{A. Proof of Hodge Decomposition Theorem}

Let $b_{ij}^\alpha = b_{ji}^\alpha = (y_{ij}^\alpha+y_{ji}^\alpha)/2$, then $y - b$ is skew-symmetric, and $\langle y-b, b\rangle=0$. So W.L.O.G, we only need to prove the theorem with skew-symmetric preference $y$.

Now, consider the following least squares problem for each $(i,j)\in E$,
\[ \bar{y}_{ij} = \arg \min_c \sum_\alpha (y^\alpha_{ij} - c)^2. \]
Define $\bar{y}\in \YY$ by $\bar{y}^\alpha_{ij} = \bar{y}_{ij}$, then define
\[ u := y - \bar{y}.\]
Clearly $u$ satisfies $\sum_\alpha u^{\alpha}_{ij}=0$ and hence $\langle u, \bar{y}\rangle=0$.

Now consider Hilbert spaces $\XX$, $\YY$, $\ZZ$ and chain map
\[ \XX \xrightarrow{D_0} \YY \xrightarrow{D_1} \ZZ \]
with the property $D_1\circ D_0=0$. Define the product Hilbert space $\H=\XX\times\YY\times\ZZ$ and let Dirac operator $\nabla: \H\to \H$ be
  \[
  \nabla = \left(
  \begin{array}{ccc}
  0 & 0 & 0 \\
  D_0 & 0 & 0 \\
  0 & D_1 & 0
  \end{array}
  \right).
  \]
Define a Laplacian operator
  $$\Delta=(\nabla+\nabla^\ast)^2=\diag(D_0^T D_0, D_0 D_0^T+D_1^T D_1, D_1 D_1^T)$$
  where $(\cdot)^T$ denotes the adjoint operator. Then by
  Rank-nullity Theorem, $\im(\nabla) + \ker(\nabla^T) = \H$, in particular the middle space admits the decomposition
  \begin{eqnarray*}
  \YY & = & \im(D_0) + \ker(D_0^T) \\
  & = & \im(D_0) + \ker(D_0^T)/\im(D_1^T) + \im(D_1^T), \\
  & & \ \ \ \ \ \ \ \mbox{since $\im(D_0)\subseteq\ker(D_1)$}, \\
  & = & \im(D_0) + \ker(D_0^T)\cap \ker(D_1) + \im(D_1^T).
  \end{eqnarray*}
  Now apply this decomposition to $\bar{y} =y-u\in \YY$, we have $D_0 x \in \im(D_0)$, $D_1^T z \in \im(D_1^T)$, and $w \in \ker(D_0^T)\cap\ker(D_1)$.

\subsection[B. Proof of Proposition]
{B. Proof of Proposition \protect \ref{prop1}}

The posterior distribution of $x$ is proportional to
\begin{eqnarray*}
&&\exp\left(-\frac{\|y - D_0x\|_2^2}{2\sigma_\epsilon^2}-\frac{\|x\|_2^2}{2\sigma_x^2}\right)\\
&=& \exp\left(-\frac{\|y - D_0x\|_2^2+\gamma\|x\|_2^2}{2\sigma_\epsilon^2}\right)\\
&\sim& \exp\left(-\frac{ (x-\mu^t)^T(L_t + \gamma I) (x-\mu^t)}{2\sigma_\epsilon^2}\right).
\end{eqnarray*}
So $x|y$ is gaussian distribution with mean $(L_t + \gamma I)^{-1}D_0^Ty$ and covariance $\sigma_\epsilon^2(L_t + \gamma I)^{-1}$.
\[y^{t+1}_{ij} = (x_i-x_j) + \epsilon_{ij}^{t+1}\]
 is a linear combination of gaussian variables, so it is also gaussian.

The KL-divergence between two gaussian distributions has an explicit formulation
\begin{eqnarray*}
&&2KL(P^{t+1}| P^{t}) \\ \nonumber
&=& (\mu^t-\mu^{t+1})^T(\sigma_\epsilon^2\Sigma^t)^{-1} (\mu^t-\mu^{t+1})\\
&&+tr((\Sigma^t)^{-1}\Sigma^{t+1}) - \ln\frac{\det(\Sigma^{t+1})}{\det(\Sigma^{t})}- n\\
&=& \frac{1}{\sigma_\epsilon^2}(\mu^t-\mu^{t+1})^T(L_t+\gamma I) (\mu^t-\mu^{t+1})  - n\\ \nonumber
&& + tr((L_t+\gamma I)(L_{t+1}+\gamma I)^{-1})\\
&& + \ln\frac{\det(L_{t+1}+\gamma I)}{\det(L_t+\gamma I)}.
\end{eqnarray*}

\subsection[C. Proof of Proposition]
{C. Proof of Proposition \protect \ref{prop2}}

Note that $\mu^t = L_{t,\gamma}^{-1}(D_0^t)^Ty^t$, so
\begin{eqnarray*}
\mu^{t+1}
&=& L_{t+1,\gamma}^{-1}(D_0^{t+1})^Ty^{t+1} \\
&=& (L_{t,\gamma}^{-1} - \frac{L_{t,\gamma}^{-1}d_{t+1}^Td_{t+1}L_{t,\gamma}^{-1}}{1 + d_{t+1}L_{t,\gamma}^{-1}d_{t+1}^T})\nonumber\\
&& \cdot((D_0^t)^Ty^t + d_{t+1}^Ty^{t+1}_{ij}) \nonumber\\
&=&\mu^{t} + \frac{y^{t+1}_{ij}-d_{t+1}\mu^t}{1+d_{t+1}L_{t,\gamma}^{-1}d_{t+1}^T} L_{t,\gamma}^{-1}d_{t+1}^T.
 \end{eqnarray*}
Moreover
\begin{eqnarray*}
 tr((L_{t,\gamma})(L_{t+1,\gamma})^{-1}) &=& tr(I - \frac{d_{t+1}^Td_{t+1}L_{t,\gamma}^{-1}}{1 + d_{t+1}L_{t,\gamma}^{-1}d_{t+1}^T})\\
 &=& n - \frac{d_{t+1}L_{t,\gamma}^{-1}d_{t+1}^T}{1 + d_{t+1}L_{t,\gamma}^{-1}d_{t+1}^T},
\end{eqnarray*}
and
\begin{eqnarray*}
 \frac{\det(L_{t+1,\gamma})}{\det(L_{t,\gamma I})} &=& \det((L_{t,\gamma})^{-1}L_{t+1,\gamma})\\
 &=& \det(I+(L_{t,\gamma})^{-1}d_{t+1}^Td_{t+1})\\
 &=& 1 - d_{t+1}(L_{t,\gamma})^{-1}d_{t+1}^T.
\end{eqnarray*}
Last equation uses $d_{t+1} = e_i-e_j$. Plugging all these identities into Proposition \ref{prop1}, we can get the result.

\end{document}